\documentclass[sigconf]{acmart} 

\usepackage{booktabs} 
\usepackage[T1]{fontenc}
\usepackage{lmodern}
\usepackage{color}

\renewcommand\footnotetextcopyrightpermission[1]{} 
\setcopyright{none}

\begin{document}
\title{Convolutional Neural Networks for Toxic Comment Classification}

\author{Spiros V.~Georgakopoulos}
\affiliation{%
  \institution{Department of Computer Science and Biomedical Informatics, University of Thessally}
  \city{Lamia}
  \country{Greece}
}
\email{spirosgeorg@uth.gr}

\author{Sotiris K.~Tasoulis}
\affiliation{%
	\institution{Department of Computer Science and Biomedical Informatics, University of Thessally}
	\city{Lamia}
	\country{Greece}
}
\email{stas@uth.gr}

\author{Aristidis G.~Vrahatis}
\affiliation{%
	\institution{Department of Computer Science and Biomedical Informatics, University of Thessally}
	\city{Lamia}
	\country{Greece}
}
\email{agvrahatis@upatras.gr}

\author{Vassilis P.~Plagianakos}
\affiliation{%
 \institution{Department of Computer Science and Biomedical Informatics, University of Thessally}
 \city{Lamia}
 \country{Greece}
}
\email{vpp@uth.gr}

\begin{abstract}

Flood of information is produced in a daily basis through the global internet usage arising from the online interactive communications among users. While this situation contributes significantly to the quality of human life, unfortunately it involves enormous dangers, since online texts with high toxicity can cause personal attacks, online harassment and bullying behaviors. This has triggered both industrial and research community in the last few years while there are several tries to identify an efficient model for online toxic comment prediction. However, these steps are still in their infancy and new approaches and frameworks are required. 
On parallel, the data explosion that appears constantly, makes the construction of new machine learning computational tools for managing this information, an imperative need. Thankfully advances in hardware, cloud computing and big data management allow the development of Deep Learning approaches appearing 
very promising performance so far.
 For text classification in particular the use of Convolutional Neural Networks (CNN) have recently been proposed approaching text analytics in a modern manner 
 emphasizing in the structure of words in a document.
 In this work, we employ this approach to discover toxic comments in a large pool of documents provided by a current Kaggle's competition regarding Wikipedia's talk page edits. To justify this decision we choose to compare CNNs against the traditional bag-of-words approach for text analysis combined with a selection of algorithms proven to be very effective in text classification. The reported results provide enough evidence that CNN enhance toxic comment classification reinforcing research interest towards this direction.


\end{abstract}

%
%


\maketitle

\section{Introduction}

Daily, we receive an avalanche of short text information from the explosion of online communication, e-commerce and the use of digital devices \cite{song2014short}. This volume of information requires text mining tools to perform a number of document operations in a timely and appropriately manner. Text classification is a classic topic for natural language processing and an essential component in many applications, such as web searching, information filtering, topic categorization and sentiment analysis \cite{aggarwal2012survey}. Text classification can be defined simply as follows: Given a set of documents D and a set of classes (or labels) C, define a function F that will assign a value from the set of C to each document in D \cite{murty2011text}. As a result, a huge pool of machine learning methodologies have been applied for text classification in various data types with satisfactory results. Nowadays, information is usually in short texts such as social networks, news in web pages, forums and so on. However, short texts collections, having the limitation of short length documents,  end up represented by sparse matrices, with minimal co-occurrence or shared context. As a result, defining efficient similarity measures is not straightforward, especially regarding the most popular word-frequency based approaches, resulting in degrading performance \cite{quan2010short}.

Recently, Convolutional Neural Networks (CNN) are being applied to text classification or natural language processing both to distributed as to discrete embedding of words \cite{dos2014deep, kim2014convolutional}, without using syntactic or semantic knowledge of a language \cite{johnson2014effective}. Indicatively, the work of Zhang et al. \cite{zhang2015character} offers an empirical study on character-level convolutional networks for text classification providing evidences that CNN using character-level features is an effective method. Also, a recurrent CNN model was proposed recently for text classification without human-designed features \cite{lai2015recurrent} by succeeding to outperform both the CNN model as well as other well-established classifiers. Their model captures contextual information with the recurrent structure and constructs the representation of text using a convolutional neural network. Meanwhile, CNN has been shown an alternative mechanism for effective use of word order for text categorization \cite{johnson2014effective}.
An effective CNN based model using word embeddings to encode texts is published recently \cite{zhang2018textual}. It uses semantic, embeddings, sentiment embeddings and lexicon embeddings for texts encoding, and three different attentions including attention vector, LSTM (Long Short Term Memory) attention and attentive pooling are integrated with CNN model. To improve the performance of three different attention CNN models, CCR (Cross-modality Consistent Regression) and transfer learning are presented. It is worth noticing that CCR and transfer learning are used in textual sentiment analysis for the first time. Finally, some experiments on two different datasets demonstrate that the proposed attention CNN models achieve the best or the next-best results against the existing state-of-the-art models.

\paragraph{Toxic Comment Discovery}

Text arising from online interactive communication hides many hazards such as fake news, online harassment and toxicity \cite{duggan2014online}. Toxic comment is not only the verbal violence but a comment that is rude, disrespectful or otherwise likely to make someone leave a discussion. Toxic comment can be considered also the personal attack, the online harassment and bullying behaviors. Unfortunately, it is a usual phenomenon in the world of web and causes several problems and with the rise of social media platforms and the explosion of online communication, the risk is increased. Indicatively, the Wikimedia foundation found that $54\%$  of those who had experienced online harassment expressed decreased participation in the particular project which occurred \cite{wulczyn2017ex}. Also, a 2014 Pew Report highlights that $73\%$ of adult internet users have seen someone harassed online, and $40\%$ have personally experienced it \cite{duggan2014online}. Although, there are efforts to enhance the safety of online environments based on crowdsourcing voting schemes or the capacity to denounce a comment, in most cases these techniques are inefficient and fail to predict a potential toxicity \cite{hosseini2017deceiving}. 

Automatic toxic comment identification and prediction in real time is of paramount importance, since it would allow the prevention of several adverse effects for internet users. Towards this direction, the work of Wulczyn et al. \cite{wulczyn2017ex} proposed a methodology that combines crowdsourcing and machine learning to analyze personal attacks at scale. Recently, Google and Jigsaw launched a project called Perspective \cite{hosseini2017deceiving}, which uses machine learning to automatically detect online insults, harassment, and abusive speech. Perspective is an API (www.perspectiveapi.com) that enables the developers to use the toxic detector running on Google's servers, to identify harassment and abuse on social media or more efficiently filtering invective from the comments on a news website. The API uses machine learning models to score the perceived impact a comment might have on a conversation. Developers and publishers can use this score to give realtime feedback to commenters or help moderators do their job, or allow readers to more easily find relevant information, as illustrated in two experiments below. 

A main limitation of these models is that they are not as reliable as it should and that usually the degree of toxicity is not determined. The latter is depicted on a Kaggle competition. Kaggle is a platform for predictive modelling and analytics competitions in which statisticians and data miners compete to produce the best models for predicting and describing the datasets uploaded by companies and users. In this work we employ the dataset provided by the current Kaggle challenge on toxic comment classification in an attempt to investigate whether the recent emergence of CNN in text mining, using word embeddings to encode texts,  provide any significant benefit over the traditional approaches for classification accompanied by  Bag-of-Words (BoW) representations.

The rest of the paper is organized as follows. In Section~\ref{sec:cnn} we present the details of the CNN's based approach for  text classification while in Section~\ref{sec:bow} we describe text prepossessing based on the Bag-of-Words model; for text analysis. Section~\ref{sec:exp} is devoted to the experimental evaluation. Finally, Section~\ref{sec:con} contains concluding remarks and pointers for future work.




\section{Convolutional Neural Networks} \label{sec:cnn}
In this section, we outline the Convolutional Neural Networks for classification and  also provide the process description for  text classification in particular.
Convolutional Neural Networks are multistage trainable Neural Networks architectures developed for classification tasks. Each of these stages, consist the types of layers described below~\cite{lecun2,fukushima}:


\begin{enumerate}
	
	\item {\it Convolutional Layers}, are major components of the CNNs. A convolutional layer consists of a number of kernel matrices that perform convolution on their input and produce an output matrix of features where a bias value is added. The learning procedures aim to train the kernel weights and biases as shared neuron connection weights.
	
	\item {\it Pooling Layers}, are also  integral components of the CNNs. The purpose of a pooling layer is to perform dimensionality reduction of the input feature images. Pooling layers make a subsampling to the output of the convolutional layer matrices combing neighboring elements. The most common pooling function is the max-pooling function, which takes the maximum value of the local neighborhoods.
	
	\item {\it Embedding Layer}, is a special component of the CNNs for text classification problems. The purpose of an embedding layer is to transform the text inputs into a suitable form for the CNN. Here, each word of a text document is transformed into a dense vector of fixed size. 
	
	\item {\it Fully-Connected Layer}, is a classic Feed-Forward Neural Network (FNN) hidden layer. It can be interpreted as a special case of the convolutional layer with kernel size $1\times1$. This type of layer belongs to the class of trainable layer weights and it is used in the final stages of CNNs.
	
\end{enumerate}

The training of CNN relies on the BackPropagation (BP) training algorithm \cite{lecun2}. The requirements of the BP algorithm is a vector with input patterns $x$ and a vector with targets $y$, respectively. The input $x_i$ is associated with the output $o_i$. Each output is compared to its corresponding desirable target and their difference provides the training error. Our goal is to find weights that minimize the cost function

\begin{equation}
\centering
E(w) = \frac{1}{n} \sum_{p=1}^{P} \sum_{j=1}^{N_L}(o_{j,p}^L - y_{j,p})^2,
\label{eq:cost_function}
\end{equation}

where $P$ the number of patterns, $o_{j,p}^L$ the output of j neuron that belongs to $L$ layer, $N_L$ the number of neurons in output layer, $y_{j,p}$ the desirable target of $j$ neuron of pattern $p$. To minimize the cost function $E(w)$, a pseudo-stochastic version of SGD algorithm, also called mini-batch Stochastic Gradient Descent (mSGD), is usually utilized~\cite{mSGD}.

\subsection{CNN for Text Classification}

The CNN have been widely applied to image classification problems due to their inner capability to exploit the two statistical properties that characterize image data, namely  `local stationarity' and `compositional structure'~\cite{CNN_graph}. Local stationarity structure can be interpreted as the attribute of an image to present dependency between neighboring pixels that is reasonably constant in local image regions. Local stationarity is exploited by the CNNs' convolution operator. 

\begin{figure}
	\centering
	\includegraphics[width = \linewidth]{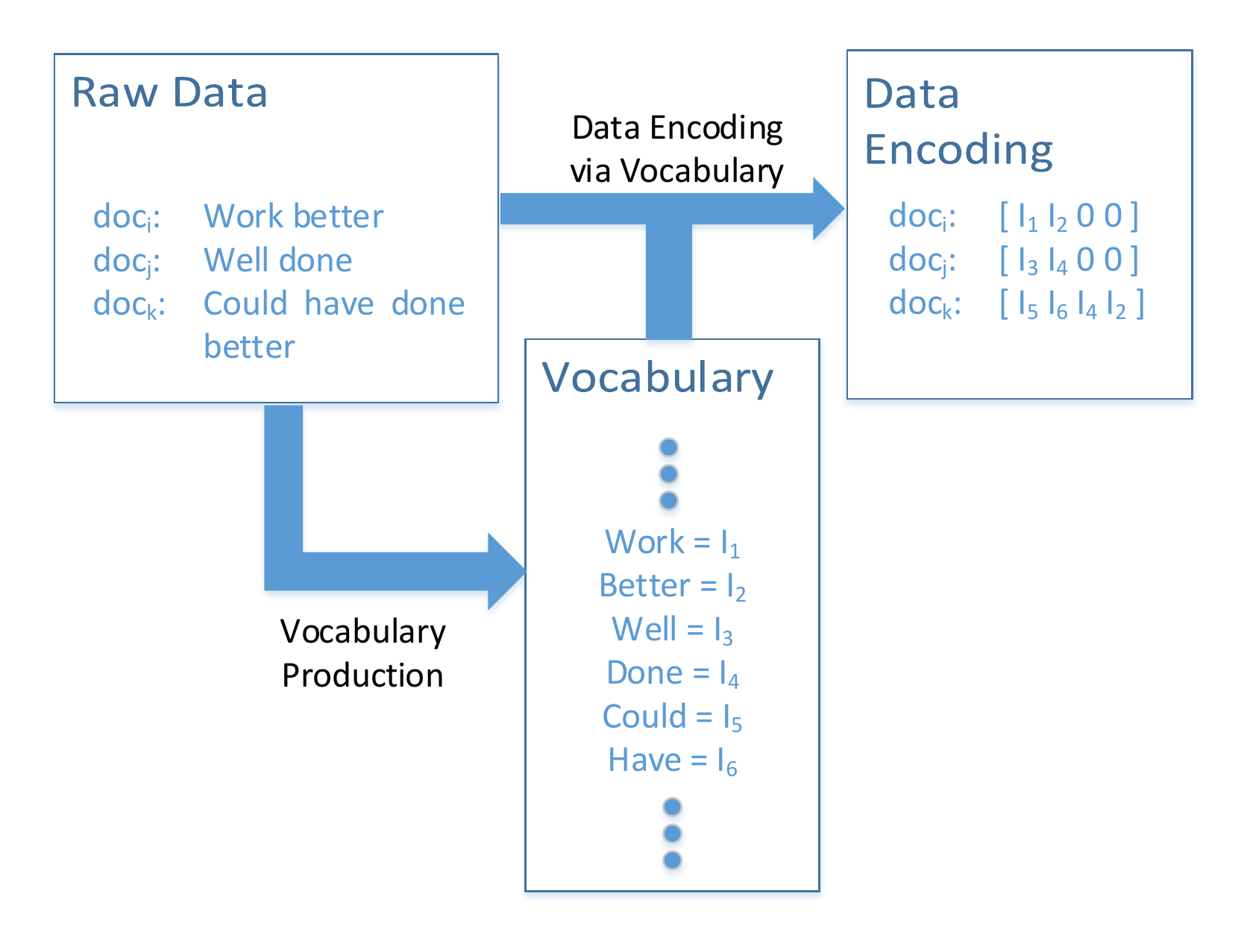}
	\caption{Example of encoding a text using a vocabulary}	\label{fig:voc}
\end{figure}

We may claim that for text classification problems the original raw data also present the aforementioned statistical properties based on the fact that neighboring words in a sentence present dependency, however, their processing is not straight forward. The components of an image are simply pixels represented by integer values within a specific range. On the other hand the components of a sentence (the words) have to be encoded before fed to the CNN~\cite{CNN_text1}.
For this purposed we may use a vocabulary.
The vocabulary is constructed as an index containing the words that appear in the set of document texts, mapping each word to an integer between $1$ and the vocabulary size. An example of this procedure is illustrated in Figure~\ref{fig:voc}. 
The variability in documents length (number of words in a document) need to be addressed as CNNs require a constant input dimensionality. For this purpose the 
padding technique is adopted, filling with zeros the document matrix in order to reach the maximum length amongst all documents in dimensionality.

\begin{figure}
	\centering
	\includegraphics[width = \linewidth]{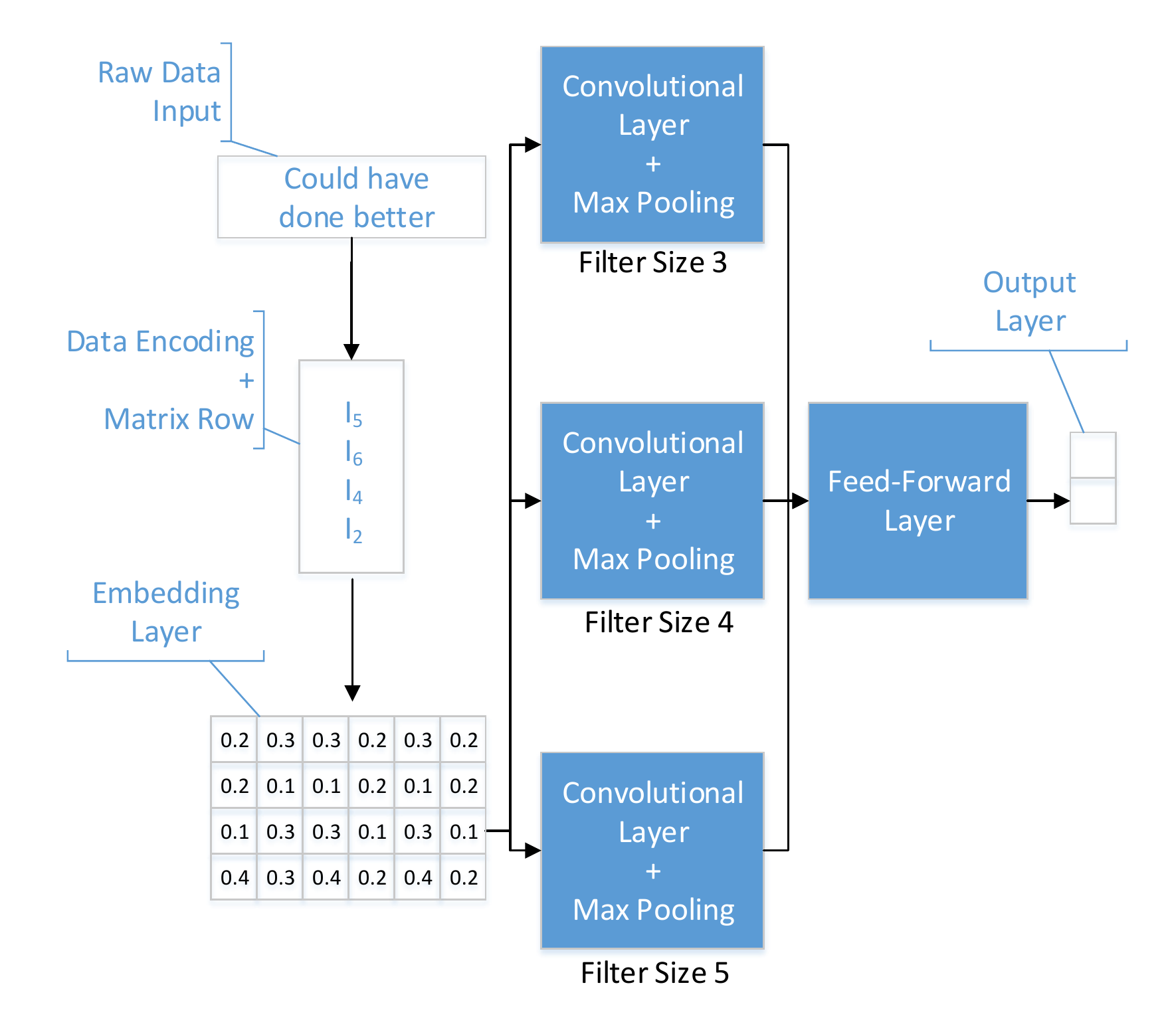}
	\caption{The CNN for text classification process.}
	\label{fig:voc2}
\end{figure}

In the next step the encoded documents are transformed into matrices for which each row corresponds to one word. The generated matrices pass through the
embedding layer where each word (row) is transformed into a low-dimension representation by a dense vector~\cite{embedding_layer}. The procedure then continues following the standard CNN methodology. At this point, it is worth mentioning that there are two approaches for the low-dimension representation of each word. The first approach called `randomized' which is achieved by placing a distribution over the word, producing a dense vector with fixed length.  The values of the vectors are tuned via the training process of the CNN. The other very popular approach also evaluated here is to employ fixed dense vectors for words, which have produced based on word embedding methods such as the word2vec~\cite{wordvec} and GloVe~\cite{glove}. In general the word embedding methods have been trained on a large volume dataset of words producing for each word a dense vector with a specific dimension and fixed values. The word2vec embedding method for example, has been trained on 100 billion words from Google News producing a vocabulary of 3 million words. The embedding layer matches the input words with the fixed dense vector of the pre-trained embedding methods that have been selected. The values of these vectors do not change during the training process, unless there are words not already included in the vocabulary of the embedding method in which case they are initialized randomly. An indicative description of the process is illustrated in Figure ~\ref{fig:voc2}.


\section{Bag-of-Words for Text Mining}\label{sec:bow}

Although word embeddings for text mining have attracted the interest of research community lately, it is still
not clear enough how beneficial can this approach be in comparison with tradition text mining approaches such as 
the BoW model. A BoW model, is an alternative way of extracting features from text that can
later be used for any kind of analysis such as classification. This representation of text describes the occurrence of words
within a document and for this purpose it only involves a vocabulary of known words and their corresponding measure of the presence.
In contrast to the model described in Section \ref{sec:cnn} any information about the order or structure of words in the document is discarded
in this case. The model is only concerned about the occurrence of words in the document and not where in the document they occur.
Despite the simplicity of this approach usually the models that use it for text categorization and classification tasks demonstrate good performance.
Although theoretical simple and practical efficient a BoW model involves several technical challenges. The first step in 
a typical text analysis procedure is to construct the  Document-Term-Matrix (DTM) from input documents.  This is done by vectorizing
documents creating a map from words to a vector space. In the next step a model for either supervised or unsupervised learning is applied. 

In what follows we provide the details of the DTM construction for the toxic comment classification problem at hand.
We begin the procedure by generating the vocabulary of words. Here we choose unique words appearing in all documents ignoring case, punctuation, numbers
and frequent words that don't contain much information (called stop words).
Then instead of scoring words based on the number of times each word appears in a document we employ the
Term Frequency - Inverse Document Frequency (TF-IDF) methodology~\cite{tfidf}.
This way we avoid the problem of high scoring by dominating words that usually do not contain `informational content'.
To achieve this the frequency of words is rescaled by how often they appear in all documents by penalizing most frequent words across all documents.
The resulting scores are a weighting indicating importance or interestingness of words. TF-IDF have proven to be
very successful for classification tasks in particular by exposing the differences amongst natural groups.
In the final step of preprocessing we deal with the sparsity problem. Sparse representations are usually more difficult to model both for computational reasons and also for information reasons, as so little information need to be extracted from such a large representational space. 
Here we choose to discard terms with higher that $99\%$ sparsity managing also to reduce the dimensionality of the DTM significantly.

\begin{figure}
	\centering
	\includegraphics[width = \linewidth]{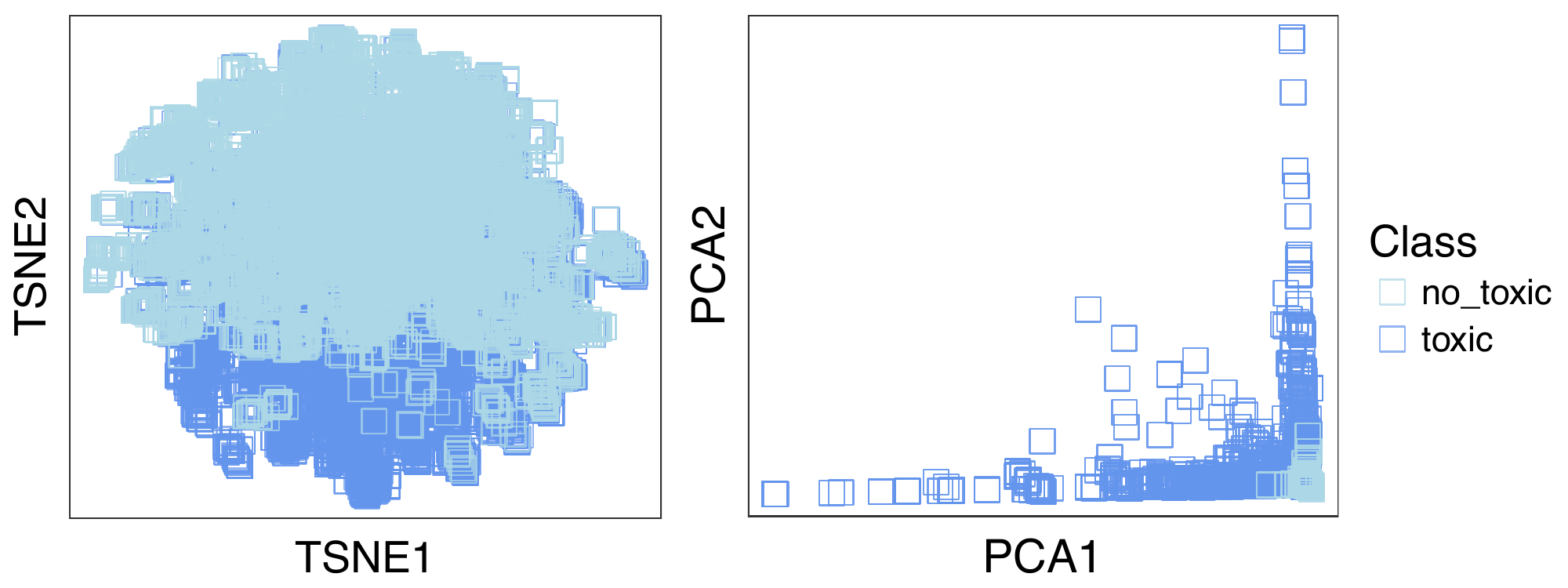}
	\caption{Two dimensional representations of the constructed DTM using PCA(right) and t-SNE(left) .}
	\label{fig:2d}
\end{figure}

In an attempt to visually validate the resulting DTM we employ two basic dimensionality reduction methods for visualizations. We employ
Principal Component Analysis~\cite{pca} to produce a two dimensional projection  and the popular t-SNE~\cite{tsne} methodology
for generating a two dimensional embedding of the  DTM (see figure \ref{fig:2d}).  For the DTM construction, samples has been selected according
to the procedure followed in the Experimental Analysis Section. In both Figures points with different color correspond to samples belonging to different classes.
We observe cluster structures in both representations indicating appropriate circumstances for a learning algorithm.
%
%
The results of applying classification methodologies on the resulting DTM are reported at the following experimental results section


\section{Experimental Analysis}\label{sec:exp}

\begin{figure*}[t]
	\centering
	\includegraphics[width = \textwidth]{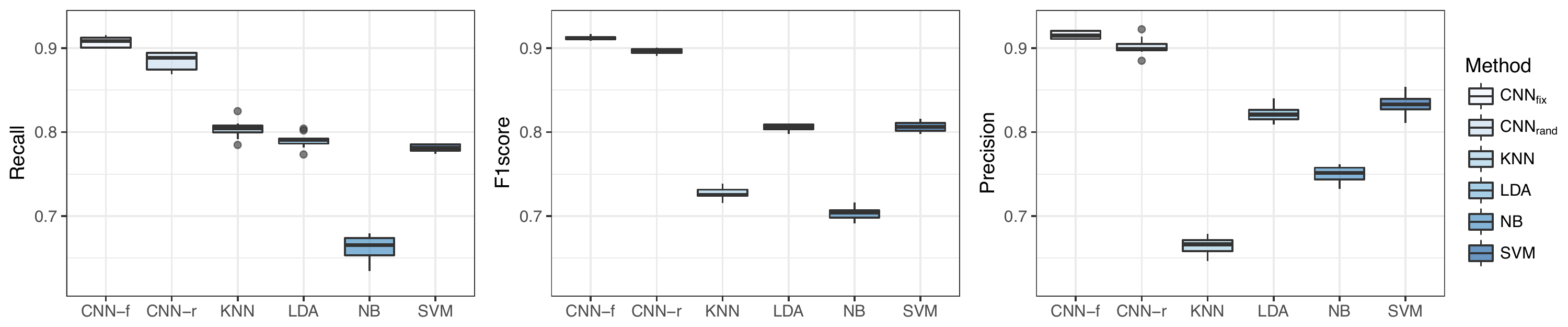}
	\caption{Box blots for  Recall, Precision and F1 score across all experiments for all Classification Methods.}
	\label{fig:res}
\end{figure*}

This section is devoted to the experimental evaluation of the presented approaches. Word embeddings and CNN are compared against the BoW approach
for which four well-established text classification methods namely  Support Vector Machines (SVM), Naive Bayes (NB), k-Nearest Neighbor (kNN) and Linear Discriminated Analysis (LDA) applied on the designed DTMs. 
We evaluate the methods for toxic comment detection employing the  dataset provide by a current Kaggle competition~\footnote{https://www.kaggle.com/c/jigsaw-toxic-comment-classification-challenge}. The dataset contain comments from Wikipedia's talk page edits which have been labeled by human raters for toxic behavior. Although there are six types of indicated toxicity {`toxic', `severe toxic', `obscene', `threat', `insult', `identity hate'} in the original dataset, all these categories were considered as toxic in order to convert our problem to binary classification.
Furthermore, for more coherent comparisons, a balanced subset of the Kaggle dataset is constructed for each evaluation of the aforementioned methods. This is achieved by random sub-sampling of the non-toxic texts, obtaining a subset with equal number of samples with the toxic texts. Subsequently,a random selection of the $80\%$ of  the resulting balanced dataset is used for training and the rest for testing. In order to provide reliable and robust inferences, each of the examined methods are evaluated $20$ times on such random separations.

The CNN architecture used here is based on the one presented in~\cite{cnn_base}. We use three different convolutional layers simultaneously, with filter size width $128$, dense vector dimension $300$. The filters width is equal to the vector dimension while their height was $3,4$ and $5$, for each convolutional layer respectively. After each convolutional layer a max-over-time pooling operation~\cite{CNN_text1} is applied. The output of the pooling layer concatenate to a fully-connected layer, while the softmax function is applied on final layer. The model is training using the SGD algorithm with mini-batches set to $64$ and learning rate $0.005$. 
Interestingly here we examine two variants of CNN approach, the $CNN_{rand}$ where the representation of the words initialized randomly and tuned during the CNN's training and the $CNN_{fix}$ where we use the pre-trained, by the word2vec model ~\footnote{https://code.google.com/archive/p/word2vec/}; low-dimensional representations of the words.
For SVM, NB and LDA we used the default parameters provided by the $R$  statistical packages {\it e1071} and {\it MASS} respectively. For kNN we choose to used a value of $k$ equal to $7$ as a representative of general performance of the method. Further parameter fitting could take place in this part of the analysis for all methods but
inevitably their performance is dominated by the construction procedure of the DTM, which has already been design considering classification performance.

Finally we perform a statistical analysis on the outcomes of the binary classification task for all methods. For this purpose we may consider, (i) samples labeled as `toxic' and predicted as `toxic' as True Positive (TP), (ii) samples labeled as `toxic' and predicted as `non-toxic' as False Negative (FN), (iii) samples labeled as `non-toxic' and predicted as `non-toxic'  as True Negative (TN) and (iv) samples labeled as `non-toxic' and predicted as `toxic' as False Positive (FP).

 A confusion matrix for each 'run' was created by calculating values in main diagnostic tests such as, True positive rate (Recall), Positive predictive value (Precision),  F1 score (see Figure \ref{fig:res}), Accuracy, False discovery rate (FDR) and True negative rate (Specificity) (see Table \ref{my-label}).
It is evident that both $CNN_{rand}$ and $CNN_{fix}$ models outperformed SVM, kNN, NB and LDA methods in all cases achieving accuracy almost over $90\%$. With respect to the toxic predictive value (Precision), CNN models had values above $90\%$ while other methods range from $65$ to $85$ percent. Almost similar behavior is observed in the True Positive Rate (Recall) and in F1-score. NB and kNN methods had the worst results having the lowest values on precision and recall respectively. This means that kNN classifies several non-toxic comments as toxic and NB classifies several toxic comments as non-toxic. Also, CNN models has the lowest variance in all cases while $CNN_{fix}$ has the best performance with respect to precision and recall. Finally we observed that CNN models have the lowest False Discovery Ratio, meaning that they mistakenly predicted as `toxic' the lowest number of `non-toxic' comments. 
The aforementioned findings state clearly that CNN  outperforms traditional text mining approaches for toxic comment classification presenting great potential for further development in toxic comment identification.

\begin{table}[h!]
	\centering
	\caption{Mean values and Standard Deviation across all experiments for Accuracy, Specificity and False discovery rate for all Classification Methods.}
	\label{my-label}
	\begin{tabular}{l l l | ll | ll}
		\toprule
		& \multicolumn{2}{c}{Accuracy} & \multicolumn{2}{c}{Specificity} & \multicolumn{2}{c}{False disc.rate} \\
		\cline{2-7}
		& Mean          & Std          & Mean           & Std            & Mean                & Std                \\
		$CNN_{fix}$   & 0.912 & 0.002 & 0.917 & 0.006 & 0.083 & 0.007 \\
		$CNN_{rand}$ & 0.895 & 0.003 & 0.906 & 0.015 & 0.092 & 0.017 \\
		kNN   & 0.697 & 0.008 & 0.590 & 0.016 & 0.335 & 0.010 \\
		LDA   & 0.808 & 0.005 & 0.826 & 0.010 & 0.179 & 0.009 \\
		NB    & 0.719 & 0.005 & 0.776 & 0.012 & 0.250 & 0.010 \\
		SVM   & 0.811 & 0.007 & 0.841 & 0.012 & 0.167 & 0.012      \\     
		\bottomrule
	\end{tabular}
\end{table}

\section{Conclusions}\label{sec:con}



Both industrial and research community in the last few years have made several tries to identify an efficient model for online toxic comment prediction
due to its importance in online interactive communications among users, but these steps are still in their infancy.
This work is devoted to the study of a recent approach for text classification involving word representations and Convolutional Neural Networks. 
We investigate its performance in comparison to more widespread text mining methodologies for the task of 
toxic comment classification.
As shown CNN can outperform well established methodologies providing enough evidence that their use is appropriate for
toxic comment  classification. The promising results are motivating for further development of CNN based methodologies for text mining in the near future, in our 
 interest, employing methods for adaptive learning and providing further comparisons with n-gram based approaches.

\section*{Acknowledgments}
We gratefully acknowledge the support of NVIDIA Corporation with the donation of the Titan X Pascal GPU used for this research.


%
%
%
%

\bibliographystyle{ACM-Reference-Format}
\bibliography{sample-bibliography,mybib,aris,newbib_spiros}

\end{document}